\documentclass[lettersize,journal]{IEEEtran}
\usepackage{amsmath,amsfonts}
\usepackage{algorithmic}
\usepackage{algorithm}
\usepackage{array}
\usepackage[caption=false,font=normalsize,labelfont=sf,textfont=sf]{subfig}
\usepackage{textcomp}
\usepackage{stfloats}
\usepackage{url}
\usepackage{verbatim}
\usepackage{comment}
\usepackage{graphicx}
\usepackage{cite}
\usepackage{enumerate}
\usepackage{url}
\usepackage{hyperref}
\usepackage{todonotes}
\usepackage{amsmath}
\usepackage{multicol}
\usepackage{graphicx}
\usepackage{booktabs}
\usepackage{caption}
\begin{document}

\onecolumn © 2024 IEEE. Personal use of this material is permitted. Permission from IEEE must be obtained for all other uses, in any current or future media, including reprinting/republishing this material for advertising or promotional purposes, creating new collective works, for resale or redistribution to servers or lists, or reuse of any copyrighted component of this work in other works.

\twocolumn
\title{Development and Evaluation of a Learning-based Model for Real-time Haptic Texture Rendering}

\author{Negin Heravi, Heather Culbertson, Allison M. Okamura, Jeannette Bohg
\thanks{N.\ Heravi was and A.\ M.\ Okamura is with the Department of Mechanical Engineering, Stanford University. H.\ Culbertson is with the Department of Computer Science, University of Southern California. J.\ Bohg is with the Department of Computer Science, Stanford University. nheravi@alumni.stanford.edu,[aokamura,bohg]@stanford.edu, hculbert@usc.edu.
}}

\markboth{Journal of \LaTeX\ Class Files,~Vol.~, No.~, Feb~2024}%
{Shell \MakeLowercase{\textit{et al.}}: A Sample Article Using IEEEtran.cls for IEEE Journals}


\maketitle

\begin{abstract}
Current Virtual Reality (VR) environments lack the haptic signals that humans experience during real-life interactions, such as the sensation of texture during lateral movement on a surface. Adding realistic haptic textures to VR environments requires a model that generalizes to variations of a user's interaction and to the wide variety of existing textures in the world. Current methodologies for haptic texture rendering exist, but they usually develop one model per texture, resulting in low scalability. We present a deep learning-based action-conditional model for haptic texture rendering and evaluate its perceptual performance in rendering realistic texture vibrations through a multi-part human user study. This model is unified over all materials and uses data from a vision-based tactile sensor (GelSight) to render the appropriate surface conditioned on the user's action in real-time. For rendering texture, we use a high-bandwidth vibrotactile transducer attached to a 3D Systems Touch device. The results of our user study shows that our learning-based method creates high-frequency texture renderings with comparable or better quality than state-of-the-art methods without the need to learn a separate model per texture. Furthermore, we show that the method is capable of rendering previously unseen textures using a single GelSight image of their surface.
\end{abstract}

\section{Introduction}

In the past decade, virtual reality (VR) has shown great promise in visually immersing users in a variety of environments due to major improvements made to realistic virtual 3D visual displays \cite{CipressoVRVisual}. However, these environments do not yet provide all the sensory signals humans experience in their physical interactions such as the sense of touch. This lack of sensory information reduces the realism of these environments, preventing them from being truly immersive, because our perception is interactive and heavily based on our expectations of how the world reacts to our actions \cite{NoeBook,Magnenatbelievability,BohgHSBKSS16}. 

Multimodal VR environments in which users can both see and feel their surroundings have immediate applications in a variety of fields such as e-commerce, training, and gaming. For example, internet shopping companies are exploring new virtual and augmented reality shopping experiences that allow customers to virtually try on products using vision \cite{AmazonVirtualTryon}. A multimodal experience would allow customers to virtually interact with products and receive real-time tactile feedback such as sensation of the texture. One could touch and feel the texture of the material of a scarf before purchasing it.

Among the missing haptic feedback signals in VR systems is the sensation of surface textures. An object’s texture is defined by its fine surface features, which can vary in both height (roughness) and spatial frequency (coarseness) \cite{Okamoto2013PsychophysicalDO}. As a user moves a hand-held probe over a real surface, a vibration signal is induced in that probe that is a function of the user's motion as well as the properties of the material they are interacting with. This vibration has been shown to affect human perception of texture \cite{KlatzkyTextureProbe}. Due to this observation about human perception, a popular method for creating virtual textures in hand-held probes is modeling the relationship between the current state of user-surface interaction and the induced vibrations in the probe \cite{Romano,RefinedARMA,Culbertson2014ModelingAR}. There have been both physics-based and data-driven approaches toward this form of haptic texture rendering, but these methods are typically not easily scalable to the large number of textures present in the real world. Physics-based approaches usually rely on complete knowledge of the object geometry and material properties, which is hard to obtain outside a lab environment. Furthermore, these models can be too computationally expensive to run in real time for haptic interactions \cite{Lin2008HapticR}.
Classical data-driven methods are capable of running in real time \cite{Romano,RefinedARMA,Culbertson2014ModelingAR}, but they typically learn a separate model per material which makes it expensive to scale these approaches to the wide variety of real materials. Also, the tools used for data-collection in these method are specialized, expensive, and not easily accessible. 

We recently proposed a deep-learning based method for haptic texture generation that is capable of learning a unified model over all materials in the training set and provided preliminary evidence regarding this model's capability to generalizing to new textures \cite{HeraviTexture}. We evaluated the performance of our model through offline (not real time) comparison of the original signal during data collection with the model output. However, the final measure of the quality of any haptic rendering algorithm should be evaluated based on its performance as judged by human users. Such evaluations depend not only on the quality of the rendering algorithm but also the capability of the accessible hardware and provide a more clear picture of the method's usability. In this paper, we build on our prior work to make the learning-based model run in real time on haptic hardware for texture rendering and conduct a multi-part human user study to evaluate its perceptual performance. The goal of our study is to assess the capability of this model for rendering high-frequency vibrations (up to 250 Hz for a rendering frequency of 500 Hz) induced in a probe as it is moved over a surface as well as its capability for rendering new materials not in its training set. Our model takes as input data from a high-resolution vision-based tactile sensor GelSight \cite{GelSight1,GelSight2} and a user's action (force and speed). The model then predicts an acceleration signal that can be directly commanded to a haptic device after accounting for the feed-forward dynamics of the actuator. We train this model using the Penn Haptic Texture Toolkit dataset \cite{Culbertson2014OneHD} augmented with GelSight images from \cite{HeraviTexture}.

The contributions of our work are as follows: 
\begin{enumerate}

\item  By concatenating human actions with a feature representation of a GelSight image of the texture, our learning-based model can predict the temporal acceleration signal in a hand-held probe during user interactions in real-time. Our model achieves this by adjusting our previously proposed generative action-conditional model that only predicts magnitude of the Discrete Fourier Transform (DFT) of the acceleration \cite{HeraviTexture}. We explain this adjustment in detail in Section \ref{section:realtimeextension}. This adjustment for real-time rendering is a key result because it is technically nontrivial and essential for interactive haptics applications.
\item Similar to our previous model\cite{HeraviTexture}, this extended model is unified across different textures, reducing the need for developing a separate model for each texture instance. 
\item Our model generalizes to previously unseen force and speed interactions as well as new instances of the modeled textures. We assess the capability of our model for generalizing to new speed and forces for haptic texture rendering in real-time in comparison to a state-of-the-art baseline using a multi-part human study.
\item We study the performance of our model along different psychophysical dimensions of texture perception: rough-smooth, hard-soft, slippery-not slippery, fine-coarse through our human study. 
\item We show that our model is capable of generalizing to unseen textures from their GelSight image for haptic texture rendering in real-time through our human study. This generalization is a unique feature of our model.
\end{enumerate}

\section{Background}

Different textures induce distinct vibratory feedback on human fingertips or a hand-held device that is moved over the surface \cite{KlatzkyTextureProbe}. The relationship between this generated vibration, the material properties of the texture, and the motion of the user is highly complex and non linear, making it difficult to model. As a result, many researchers have modeled this relationship in a data-driven fashion using,  e.g., a piece-wise auto-regressive model \cite{Culbertson2014OneHD}, a model based on the auto-regressive moving average (ARMA) \cite{Culbertson2014ModelingAR}, frequency decomposed neural networks \cite{FreqDecomposedNNTextureRendering}, waveform segment tables \cite{WaveformTexture}, and deep spatio-temporal networks \cite{SpationTemporalTextureRendering}. All these prior works fit one model per texture and do not learn the relationship between different materials. For example, \cite{Culbertson2014OneHD} used piece-wise auto-regressive models to model 100 different textures, but given a never-seen-before texture, this approach requires at least 10 seconds of data collection using a specialized device before model is fit and rendered -- even if a very similar texture already exists in the database. This makes such models hard to scale to the large number of textures available in the world because they would require (1) additional expensive data collection every time a new texture is to be added to the database and (2) saving and maintaining a large database of models for all the desired virtual textures. More recently, \cite{ShihanHeatherPrefernce} built on the piece wise auto-regressive models of \cite{Culbertson2014ModelingAR} by using a GAN-based texture model generator that refines these models using input from a human user about their preferences. They showed that this preference-driven interactive method matches or exceeds the realism of other state-of-the-art data driven methods for haptic texture rendering, but they rely on feedback from a human user to achieve this performance. 

Unlike haptic interaction datasets that are expensive and require specialized hardware to collect, visual or vision-based tactile texture data mainly require access to a camera and relatively inexpensive hardware. As a result, one can mitigate the challenge of scalability if one maps the relationship between visual information and vibro-tactile surface properties. However, the relationship between visual and haptic data is difficult to model, especially for materials and objects the model has not seen before. Others \cite{ConstrainedLMT,Burka,LMTMulti} have used visual information alone or in combination with other modalities such as sound, acceleration, and friction to learn representations for texture classification purposes, but they did not study the utility of these representations for texture rendering. Ujitoko and Ban \cite{Ujitoko2018VibrotactileSG} used Generative Adversial Networks \cite{GANs} for texture generation from images, but their method is only applicable for rendering vibratory feedback for predefined and constrained user actions and is not capable of rendering a truly interactive surface where the users can freely move.  

In our prior work \cite{HeraviTexture}, we showed that a deep-learning-based action-conditional model can learn the relationship between the vibratory signal induced in a probe as it is moved over a surface given the GelSight\cite{GelSight1,GelSight2} image of that surface and the user's action as input. Furthermore, we showed preliminary results about the capability of our model to generalize to not-seen-before textures using their GelSight image. Our model achieved this by learning a unified model that connects and learns the relationship between different textures in its texture representation space. The output of our model was the magnitude of the Discrete Fourier Transform (DFT) of the desired acceleration in the probe. We used these DFTs to reconstruct the acceleration signal with an offline algorithm. However, rendering haptic textures on a device requires reconstruction of the desired temporal vibratory signal in real time. As a result, we were only able to evaluate the performance of our output by comparing our prediction to the ground truth signal in offline data using Euclidean distance as metric. This metric did not provide a perceptual evaluation of our system. In this paper, we build on our prior work to create a model capable of rendering haptic textures in real-time using our previous learning-based generative model that predicts DFTs followed by an online real-time phase-retrieval algorithm. This enables us to evaluate the performance of our model through a human user study.

\begin{figure*}[ht]
\begin{center}
\includegraphics[width=\linewidth]{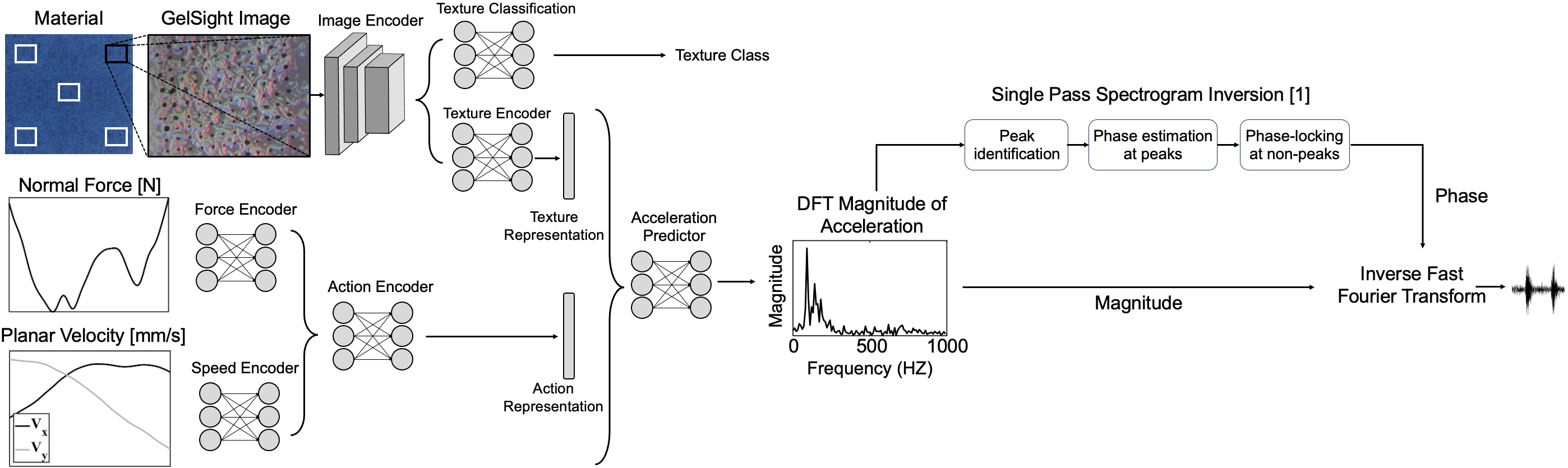}
\end{center}
\caption{High-level overview of our learning-based real-time texture generation model. First, a generative action-conditional model takes as input a GelSight image of the material as well as the user's force and planar velocity as input. An image encoder followed by a texture encoder processes the texture information from the GelSight image into a texture representation vector while a force, speed, and action encoder processe the user's force and velocity information into an action representation. The acceleration predictor module then outputs the magnitude of the spectral content of the the generated acceleration induced on this material due to this action. The model is trained in two stages. We first train the image encoder followed by a classification module for the proxy task of texture classification, and then use this encoder's frozen weights in training of the rest of the modules. During inference, we use the predicted DFT magnitude of acceleration from our generative model to construct the temporally rendered acceleration signal using the Single Pass Spectrogram Inversion (SPSI\cite{SPSI}) algorithm.}

\label{fig:overview}
\end{figure*}

\section{Texture Modeling and Rendering Methods}
We use a deep learning-based generative action conditional model based on our prior work \cite{HeraviTexture} for texture generation that we build upon for real-time haptic texture rendering. Section \ref{TextureGenerationMethod} provides a brief summary of our original method for texture generation, and the adjustments we needed to make for it to be used as a component in a real-time setting. Section \ref{section:realtimeextension} describes the extended module which is a new addition to our previous framework that allows real-time rendering. Figure \ref{fig:overview} shows a high-level overview of the entire model.

\subsection{Learning-based Texture Generation}
\label{TextureGenerationMethod}
 In this section, we summarize the learning-based texture generation module and refer the reader to \cite{HeraviTexture} for additional details about the neural network architecture and training. 
 
We model the vibratory feedback generated in a hand-held probe as it moves over a surface. The generated vibratory feedback is a function of the surface we are moving over as well as the user's action. Our model takes as input a single image from a vision-based high-resolution tactile sensor GelSight as well as the force and speed of the user's interaction for a window of time and outputs the magnitude of the discrete Fourier transform (DFT) of the generated acceleration in the probe for a horizon of 0.1 seconds. This magnitude DFT is used as input to the temporal signal reconstruction module explained in Section \ref{section:realtimeextension} to construct a final acceleration signal that can be used to actuate the device in real-time. 

In this formulation, the GelSight image provides information about the texture we are interacting with while the force and speed information indicate the user’s action. The time window size of the force and speed input should be chosen based on the sampling rate during rendering. We empirically chose a window size of 2 milliseconds for our experiments (10 data points as input at loop-rate of 500 Hz). If the model is to be used at a new rendering frequency, force and speed should either be sampled at 500 Hz or their respective encoder sizes should be adjusted, and the network should be retrained. As the Single Pass Spectrogram Inversion (SPSI) algorithm \cite{SPSI} used in the following step assumes the DFT input to be windowed using a Hanning window, our trained model predicts the magnitude DFTs with the same windowing algorithm. We used the Euclidean distance between the predicted and the ground truth magnitude up to 1000 Hz (100 DFT bins) as the training loss. 

Our model predicts only the magnitude portion of the Fourier transform, motivated by evidence that human tactile perception for texture is invariant to phase shifts \cite{KlatzkyTextureProbe}. Furthermore, we were inspired by the literature in the neural network-based audio generation community that predict the signal in the frequency domain followed by a wave reconstruction algorithm instead of predicting the wave form directly \cite{Wang2017TacotronTE}. 

For the model architecture, we used a convolutional neural network similar to Alexnet\cite{Alexnet} to encode the information in the GelSight image. Fully connected layers with rectified linear units in between were used for texture classification (layer sizes: 8960,4096,512,93), force (10,10,10,10), speed (20,20,20,20,10), action (20,400,300,200,100) and texture encoders (8960,4096,512,256) as well as the acceleration prediction (356,300,300,200,100,100) network. The architecture is trained in two stages where the image encoder and the texture classification layers are trained for a classification task. Afterwards, we freeze the weights of the image encoder, and train the network for the DFT prediction task. The sizes of the fully connected layers and the choice of freezing the pre-trained weights were done empirically. We refer the reader to \cite{HeraviTexture} for more details. 

During inference, for materials in the training set, we extract the texture representation vector from their GelSight image using the trained image encoder and use that representation vector directly as input to the acceleration prediction module. For rendering materials not in the training set, we encode their GelSight image to extract a representation vector in a similar fashion as before, but then find and use the nearest neighbor of this vector in the training set as texture representation input to the acceleration prediction module. Because our neural network is trained on texture representations for less than 100 textures, it cannot always smoothly extrapolate to new data points in the high dimensional (256) texture representation space. This can result in non-optimal noisy outputs from the acceleration prediction network as it has not been trained on these new texture representation vectors. We chose to use the nearest neighbor representation instead of the raw vector to address this issue. Our preliminary tests showed this choice results in higher quality virtual textures. Given the texture representation, a forward pass through the rest of the network is calculated at each rendering loop. Force and speed matrices are set to 0 at the beginning of the rendering and updated at every loop. We used a speed threshold of 5 millimeter per second where we only played the predicted signal to the user if the speed value was higher than this threshold. 

\subsection{Real-time Temporal Signal Construction from Magnitude DFTs}
\label{section:realtimeextension}
Rendering haptic textures requires commanding a temporal acceleration signal to a vibrotactile actuator. However, the output of the learning-based texture generation model \cite{HeraviTexture} as reviewed in Section \ref{TextureGenerationMethod}  is a short-term DFT magnitude for 0.1 seconds. To use this model for real-time haptic texture rendering, we build on our prior work by combining these short-term DFT magnitudes to reconstruct a temporal acceleration signal in real-time. To achieve this, as the user moves across a surface, we continuously use their force and speed measurements as input to the texture generation model to create a running estimate of the short-term DFT magnitude of the output acceleration over time with overlapping time windows. As an example, at time \textit{t}, for a rendering algorithm running at a frequency of 500 Hz, the model predicts the DFT for a window of [\textit{t}, \textit{t+0.1}]. In the following loop of the algorithm at time \textit{t+0.002} we are predicting the DFT for a window of [\textit{t+0.002},\textit{t+0.102}]. 

Afterward, we use the redundancy in the information caused by the overlapping in the window of the magnitude DFTs to retrieve the phase component. Since the haptic rendering needs to run online in real time, the phase retrieval algorithm chosen for this step can only depend on information from the past. This constraint prevents us from using common phase reconstruction algorithms such as by Griffin and Lim \cite{GLA} because the proposed algorithm is an iterative method where the prediction for each frame depends on all future and past frames. Instead, we use the Single Pass Spectrogram Inversion (SPSI) \cite{SPSI} algorithm which is a non-iterative, fast method that provides phase estimates only based on previously observed DFTs. This approach assumes the signal we are trying to reconstruct is composed of a series of sinusoidal periodic waves with slowly varying amplitudes. This means that for a series of consecutive short-term magnitude DFTs with overlapping windows, the changes in neighboring magnitude DFTs should be due to the phase component of the signal we are trying to reconstruct. The algorithm estimates this phase component by first randomly initializing a phase estimate. Afterwards, it works by finding the frequency bins corresponding to peaks in the magnitude spectrogram. It then estimates the phase at these peaks by conducting a linear extrapolation based on the frequency value of this bin and its value at the previous time-steps \cite{SPSI}. The phase for all non-peak bins are then assigned based on the logic in phase-locked vocoders described in \cite{phaselockedvocoders}. We used this algorithm to reconstruct the phase values from the stream of magnitude DFTs.

Given the magnitude and reconstructed phase values, we used an inverse Fourier transform function to obtain a running estimate of the temporal signal that we command to the device at a loop rate frequency of 500 Hz after accounting for the feed-forward dynamics of the actuator. We chose this loop rate frequency because the computation time for forward passing through our network and the SPSI optimization was about 0.002 seconds.

\begin{figure*}[ht]
\begin{center}
\includegraphics[width=1.0\linewidth]{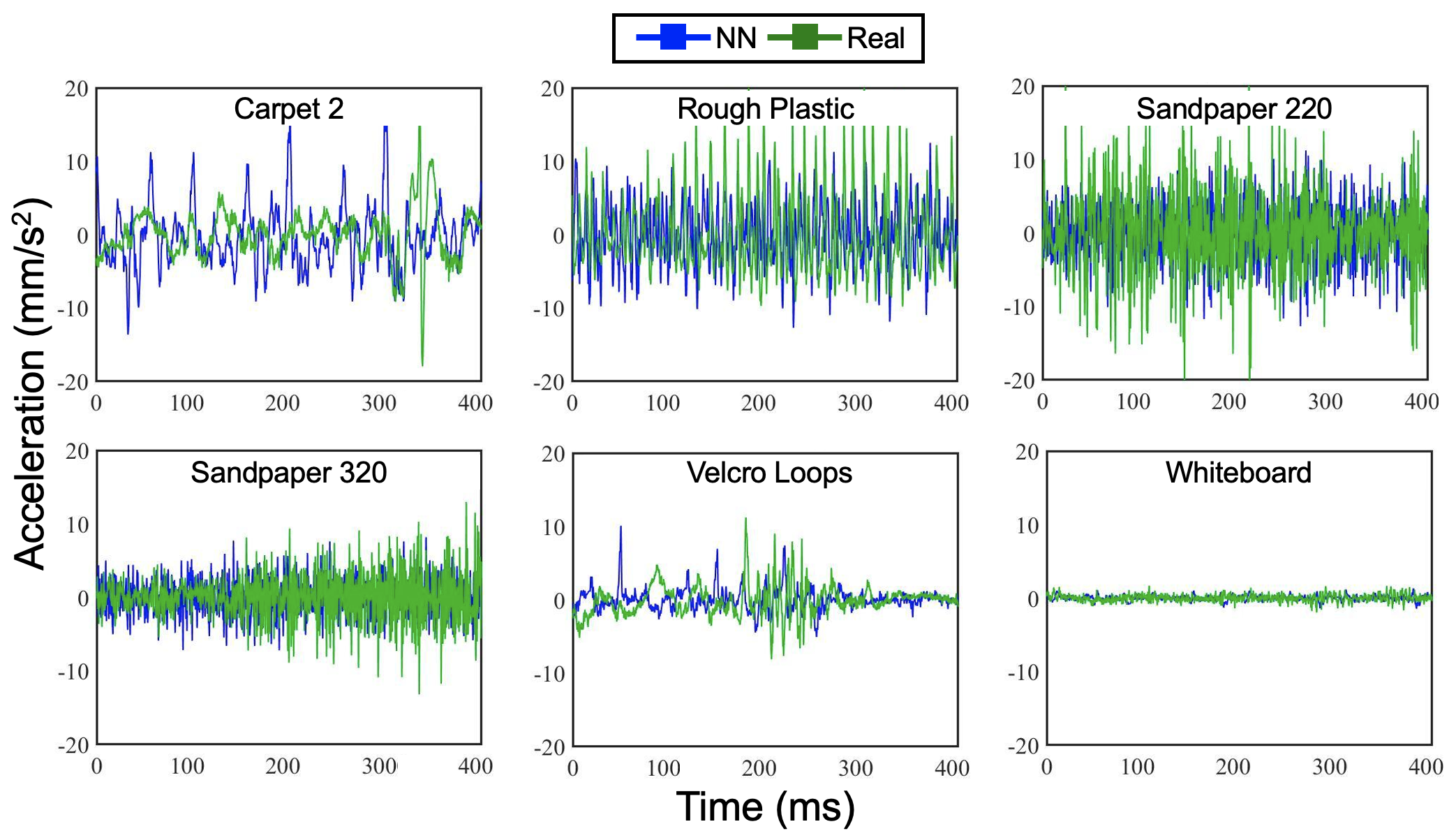}
\end{center}
\caption{Comparison of our generated acceleration signal and the ground truth recordings for a few materials in the HaTT dataset. NN is the predicted generation of our neural network-based model while Real is the ground truth acceleration readings.}

\label{fig:samplerender}
\end{figure*}

\subsection{Rendering Setup}
\label{haptuatorhardware}
For our human user study described in detail in Section \ref{userstudy}, the participants explored different surfaces using a hand-held probe in real and virtual conditions. Virtual textures were rendered using a hand-held force feedback haptic device, the 3D Systems Touch, augmented with a high-bandwidth vibrotactile transducer, the Haptuator (HapCoil-One model), attached to the handle. The Haptuator was firmly mounted and aligned axially with the handle near the tip using zipties and was commanded using a SENSORAY 626 PCI card followed by a custom-built linear current amplifier via an LM675T op-amp with a gain of 1 A/V at a rate of 500 Hz. We used a feed-forward dynamic compensation controller similar to \cite{DyanmicModelingHaptuator} to account for the Haptuator dynamics. During virtual rendering, we use the position measurements of the Touch device to calculate the velocity of the user during motion. To reduce the effect of noise on velocity readings, we low-pass filtered the velocity signals at 20 Hz similar to \cite{Culbertson2014OneHD}. For force feedback calculations, we used Hooke's law with a gain of 0.5 N/mm multiplied by the displacement $\vec{\delta}$ calculated as the penetration of the device into the virtual floor. Assuming the coordinate frame is chosen such that the virtual floor plane's normal is in the z direction, the magnitude of the normal force applied to the plane is equal to the magnitude of the z component of the force feedback:
\begin{equation}
\begin{split}
\begin{gathered}
\vec{F}_{feedback} = 0.5   \vec{\delta} \\
\vec{\delta} = ( {\delta}_{x},{\delta}_{y},{\delta}_{z}) \\
\| \vec{F}_{normal} \| =  \| 0.5   {\delta}_{z} \| \\
\end{gathered}
\end{split}
\end{equation}

We used this calculated  normal force magnitude and the planar projection of the filtered velocity vector on the floor ($\dot{\delta}_{x},\dot {\delta}_{y}$) as input to our model $\phi$ for texture vibration rendering:

\begin{equation}
\begin{split}
\begin{gathered}
 Acceleration_{DFT} =\phi(\| \vec{F}_{normal} \|,\dot{\delta}_{x},\dot{\delta}_{y})
\end{gathered}
\end{split}
 \end{equation}

The Touch handle provided the force feedback while the Haptuator played the texture vibration.  Figure \ref{fig:samplerender} shows examples of the generated acceleration signal for a few materials compared to the ground truth recordings from the HaTT dataset \cite{Culbertson2014OneHD}. Because the 3D Touch device has a pen shape structure and mass similar to the data collection pen used in the HaTT dataset, and the Haptuator is mounted at around the same height as the embedded accelerometer in the data collection pen, we chose to directly command this acceleration signal to the Haptuator after accounting for the Haptuator's feed-forward dynamics using a similar method to \cite{Culbertson2014ModelingAR}. However, one would need to adjust these predicted values if they were to apply the  model to a device with a different form factor or material than the one used during data collection.

\section{Perception Experimental Methods}
\label{userstudy}
We had conducted an extensive numerical analysis of our method compared to the baseline and the ground truth signal in our previous work \cite{HeraviTexture}.  In this paper, we extend that analysis by conducting a user study to assess the perceptual effectiveness of our machine learning-based haptic texturing rendering method. The study had three objectives: 
\begin{enumerate}
\item Measure the capability of our method in rendering realistic textures compared to that of the piece-wise AR baseline. 
\item Study the underlying perceptual dimension that differentiates the real material and the two virtual rendering methods (ours, piece-wise AR baseline) by asking the participants to rate each texture along different adjective axes. 
\item Assess the capabilities of our method to generalize to rendering texture sensation of new materials given their GelSight image. \end{enumerate}
The study protocol was approved by the Stanford Institutional Review Board, and all participants gave informed consent. 25 people (10 females and 15 males) in the age range of 21-30 were recruited for this study. All the participants were right-handed. Among the participants, 6 people reported no prior experience with human-machine interactive devices, and  8, 4, and 7 people reported limited, moderate, and extensive prior experience levels, respectively.

\begin{figure*}[ht!]
\begin{center}
\includegraphics[width=1.0\linewidth]{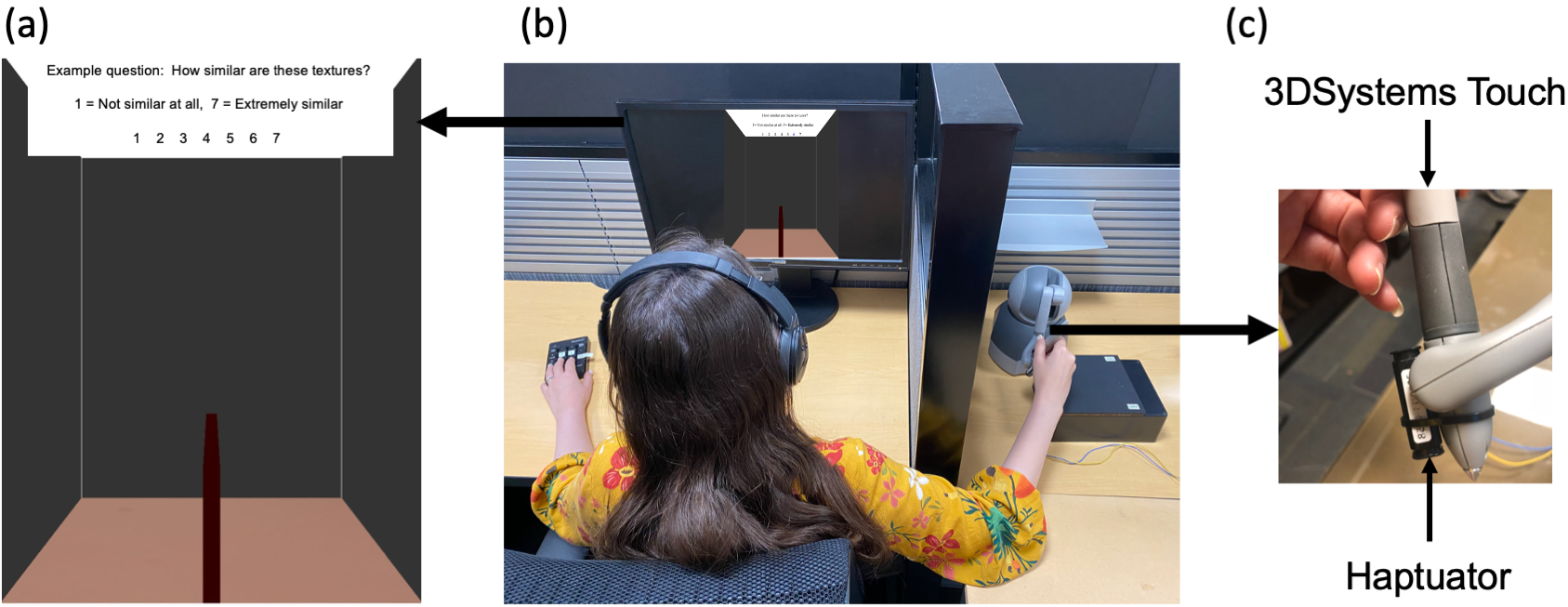}
\end{center}
\caption{Experimental setup. Participants sat at a table in front of a monitor that displayed the virtual environment and held a 3D systems Touch stylus augmented with a Haptuator behind a divider to their right. The divider blocked the participant's view of their hands and the real textures. Using this setup participants interacted with virtual (on left side on the Touch device here) and/or real (on right side of the Touch device here) surfaces and answered questions shown on the screen using a keyboard. To mask the sound of the device and the environment participants wore headphones playing white noise.}
\label{fig:haptuatorsetup}
\end{figure*}

\subsection{User Study Setup}
The participants sat in front of a monitor that displayed the virtual environment as shown in Figure \ref{fig:haptuatorsetup}b. To avoid visual cues affecting the texture sensation, all the texture surfaces were shown as a solid light-brown color on the screen as shown in Figure \ref{fig:haptuatorsetup}a. A table was placed to the right of the participants where they held the stylus of a 3D Systems Touch device augmented with a Haptuator to interact with virtual surfaces as explained in Section \ref{haptuatorhardware} and shown in Figure \ref{fig:haptuatorsetup}c. The participants also interacted with real surfaces that were mounted on a piece of acrylic using double sided tape and were raised to be at the same height of the virtual textures using the same tool. To avoid the tip of the Touch device getting stuck in the real materials such as carpet, a spherical tip was mounted on the tip. The Haptuator was not activated during real texture explorations. Our setup was inspired by the setup in \cite{HeatherDoesForceMatter}.

Because the focus of this study was assessing the capability of our model in rendering the texture vibration, we set stiffness to a constant values of 0.5 Newtons per millimeter and friction to 0 (no friction beyond the device's natural friction) for all parts of this study. A divider was placed between the participant and their hand to block the view of their hand's motion and the real material they were interacting with when a real material was present. The participants placed their hand behind the divider and held the Touch handle in order to explore surfaces. Figure \ref{fig:haptuatorsetup}b shows a general overview of our experimental setup. 

To mask the sound of the device and noise in the testing environment, the participants wore headphones that played white noise during the study. We instructed all participants to rest their elbow on the table to their right during exploration and only move their forearm and palm to explore the textures using the handle. We also instructed the subjects to avoid tapping on the surfaces and to bring down the pen slowly upon contact initialization to avoid accidental tapping upon contact. A wait/explore sign on the screen controlled by the experimenter indicated to the participant when the surfaces were ready to be touched. Participants used their left hand to type their answers to questions or to press next when they had finished surface exploration.

\begin{figure*}[ht]
\begin{center}
\includegraphics[width=1.0\linewidth]{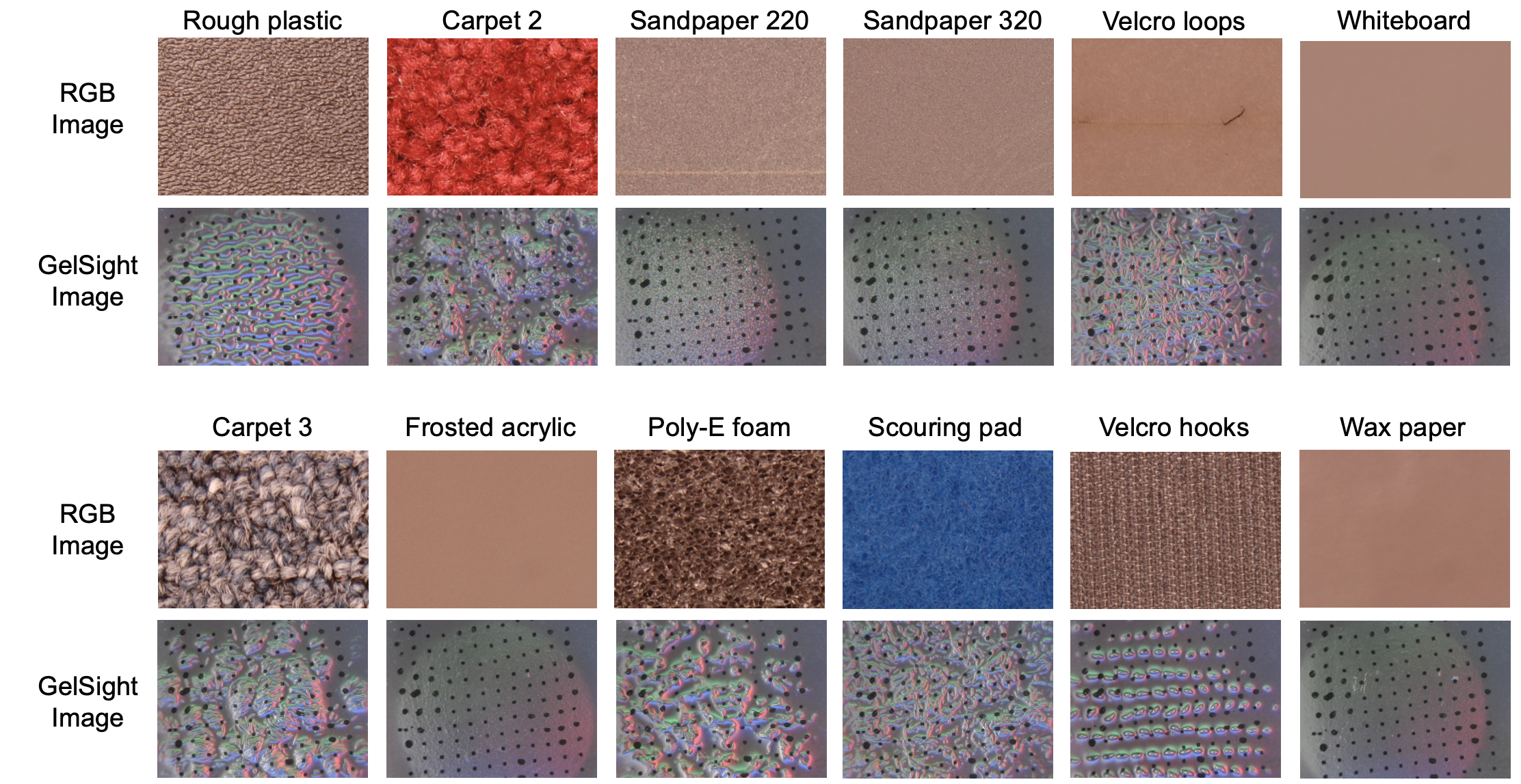}
\end{center}
\caption{RGB and GelSight images of the 12 textures from the Penn Haptic Texture Toolkit\cite{Culbertson2014OneHD} used and modeled in our user study. RGB images from \cite{Culbertson2014OneHD}.}

\label{fig:hapticstudytextures}
\end{figure*}

\subsection{Procedure}
The first step in the procedure was training, in which participants freely explored real textures placed on the right side of the Touch device on the table with the handle without the Haptuator being activated to familiarize themselves with texture sensation through the handle and the setup. 12 textures were presented on the right side of the workspace for exploration in random order. There were no questions asked in this part of the study, and participants pressed a button labeled with ``Next'' on a keyboard to move on to the next texture once they had finished exploring. These textures were close replicas of those used in the HaTT dataset. 12 textures were chosen to cover a wide range of characteristics from rough to smooth while keeping the study session within reasonable length. 

The main study was conducted in four parts: similarity comparison, forced-choice comparison, adjective rating, and generalization to rendering novel unseen textures. We used the real and virtual versions of the 12 textures shown in Figure \ref{fig:hapticstudytextures} for the first two parts of this study, and the first 6 out of the 12 textures for the third part. The last part of the study used only the virtual version of the first 6 textures as well as the virtual version of the novel unseen textures. The study parts are as follows:

\subsubsection{Similarity Comparison}
In the first part of the study, the participants were exploring both sides of the screen and rating the similarity of the two sides compared to each other using integer values on a 1 to 7 Likert scale (1: not similar at all, 7: extremely similar).  Each participant answered 36 questions covering all three pairs of (piece-wise AR baseline, real material), (our method, real material), and (our method, piece-wise AR baseline) for each of the textures in a random order. For each pair, we randomized the side where each model was displayed per question. 

\subsubsection{Forced-choice Comparison}
The participants first explored a reference real texture on the right side of the divider. Then, they explored two virtual versions of the texture (piece-wise AR and ours) on both sides and were asked to choose the side that is closest to the reference texture they just experienced. This allowed us to obtain a direct comparison of the two virtual rendering methods because the participants were forced to choose one versus the other. We hypothesize that our model has equal or better performance compared to the piece-wise AR baseline in recreating the real textures, and hence participants will choose it as the closer option in this part of the study.

\subsubsection{Adjective Ratings}
In the adjective rating experiments, participants interacted with textures one at a time on the right side of the screen, and they were rating each texture on 4 different scales of rough-smooth, hard-soft, slipper-no slippery, and fine-coarse similar to \cite{Culbertson2014ModelingAR}. These axes have been shown to cover the psychophysical dimensions of tool-mediated texture perception \cite{Okamoto2013PsychophysicalDO}. The scales for each adjective dimension were integer values of 1 to 7, and no description about the meaning of each adjective was provided to the participant. As the goal of our experiments in this paper is to assess the capability of our rendering method for modeling vibrations due to interactions, the important axis to study are rough-smoothness and fine-coarseness, but we included the slipperiness and hardness axis as well for the sake of completeness. We hypothesize that our model will be rated at the same distance or closer to the real material than the AR model along the roughness and fineness axes. 

\subsubsection{Generalization to Rendering Unseen Textures}
For the final part, we were interested in assessing our model's capability to generalize to unseen textures using their GelSight image. Participants were asked to rate the similarity of the virtual model of these unseen textures compared to the virtual model of the 6 base materials used in the previous part of the experiment on a scale of 1 to 7. Each participant was asked 24 similarity rating questions for all possible pairs of 4 unseen materials versus 6 base materials. When our model performs well at generalizing to these new unseen textures, they are placed close to similarly feeling materials in the latent representation space resulting in a similar virtual texture being rendered. As a result, our hypothesis is that the participants will rate these novel virtual textures as the most similar to textures with the same class or texture properties among the 6 base materials. For example, if this hypothesis is true, we would expect a new carpet to be rated more similar to a carpet in the base materials and further from a smooth surfaces such as Whiteboard. Because there were no real textures in this part of the study, there were no wait/explore prompt signs on the screen, and the participants were allowed to move through the questions at their own pace.

After the user study, a post-survey asked the participants to rate the difficulty of each study part on a scale of 1 (very easy) to 5 (very hard). In addition, participants optionally reported strategies that they used during exploration.  

\subsection{Metrics}
We compared distributions across different conditions first using a non-parametric repeated measure Friedman test followed by post-hoc Wilcoxon tests with continuity and Bonferroni corrections where appropriate. We used Pingouin \cite{Vallat2018}, an open-source statistical package in Python, for our analysis. The reason for choosing non-parametric tests instead of ANOVA was that some of our data groups failed the normalization assumptions of ANOVA using a Skewtest function from SciPy's statistics library \cite{2020SciPy-NMeth}. Studies are inconclusive about the appropriateness of usage of ANOVA for Likert data when data is skewed \cite{ANOVALikert}, and it is suggested to use non-parametric tests in such cases. We used a significance p-value threshold of 0.05 with Bonferroni correction (threshold divided by the number of tests in each section) for statistical significance.

\section{Perception Experiments Results and Discussion}
\subsubsection{Similarity Comparison}
\begin{figure*}[ht]
\begin{center}
\includegraphics[width=1.0\linewidth]{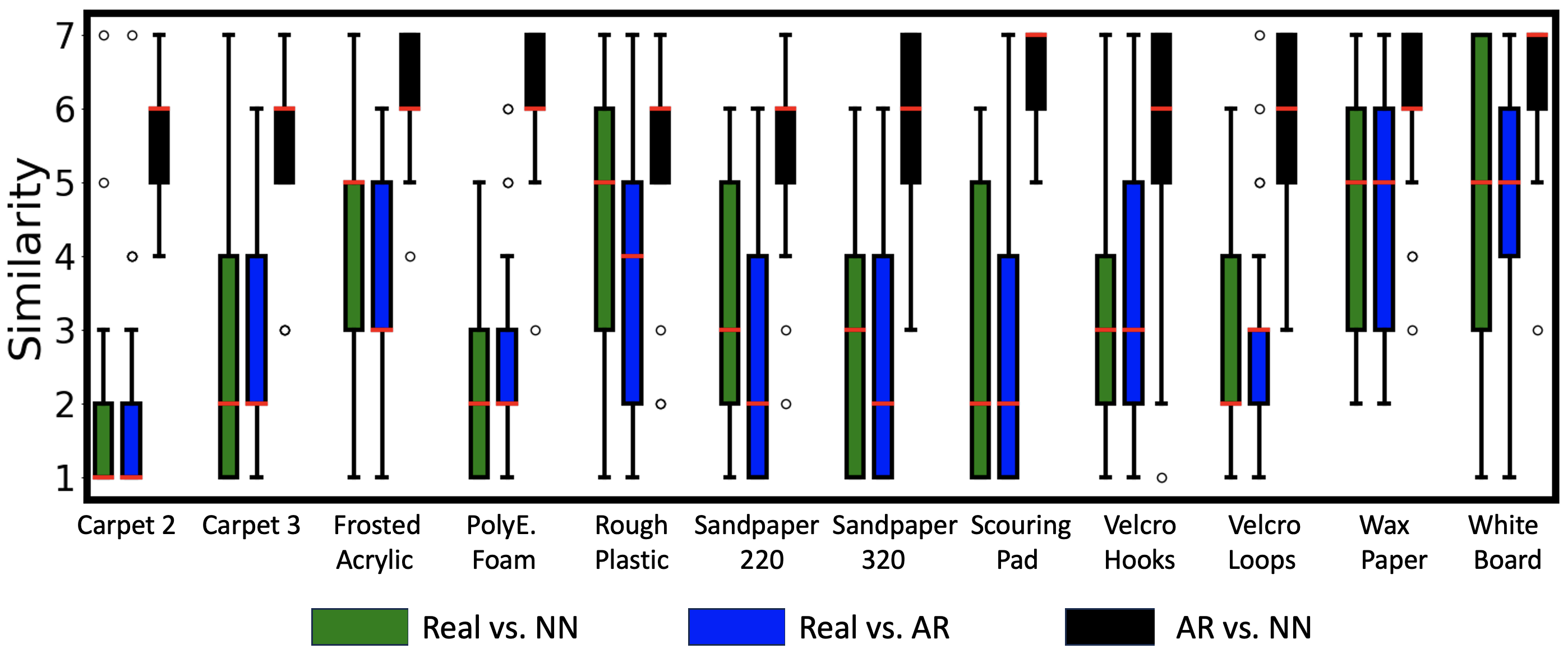}
\end{center}
\caption{Similarity rating comparisons between our proposed method (NN), piece-wise Auto Regressive baseline (AR), and the real materials (Real). The box shows the range from the first to the third quartile of the ratings across all participants. The red lines indicate the medians. The whiskers extend the box by 1.5 of the inter-quartile range (IQR) as defined by Tukey's definition of boxplots. Our method shows similarity scores close to that of the piece-wise AR baseline during rendering. We make this observation based on the comparison of Real vs.\ NN and Real vs.\ AR scores. Using a two-sided Wilcoxon test, we cannot conclude that statistically significantly difference exists between our model and the baseline AR model. The virtual textures are rated more similar to each other than the real material as expected. This is due to the fact that the virtual models only model acceleration and do not capture other aspects of the real texture. Using a Bonferroni adjusted significance level of 0.002, we found the distributions of similarity ratings of Real vs.\ NN to be statistically significantly different than those of AR vs.\ NN for all materials except Rough Plastic, Wax paper, and Whiteboard.}
\label{fig:similarityexp}
\end{figure*}

Figure \ref{fig:similarityexp} shows the results for the similarity comparison experiment per material. We evaluate the performance of the two virtual rendering methods (NN and AR) by comparing the participants' ratings for these models' similarity to the real texture. Our hypothesis is that similarity rating values for Real vs.\ NN comparisons come from the same distribution as those in Real vs.\ AR. Failure to reject this hypothesis would suggest that we cannot conclude a significant difference exists between the two models when compared to real texture. A Friedman test for each texture using the model pair-type as factor (Real vs.\ NN, Real vs.\ AR, and AR vs.\ NN) showed statistical significance for all textures. However, post-hoc two-sided Wilcoxon tests with a Bonferroni-corrected significance cut-off of 0.002 did not show a statistically significant different performance between our model and that of the piece-wise AR baseline across all materials. Table \ref{pvaluetable} row 1 shows the p-values for these Wilcoxon tests. This data does not reject our initial hypothesis that the Real vs.\ NN and Real vs.\ AR ratings are similar. It should be noted that our model is achieving this performance \emph{without} the need for learning a separate model for each material as is required for AR.

\begin{table*}[bt]
\centering
\footnotesize
\scalebox{0.9}{\begin{tabular}{cccccccccccccc}
\multicolumn{1}{c}{\textbf{}} & 
\multicolumn{1}{c}{\textbf{Carpet}} &       \multicolumn{1}{c}{\textbf{Carpet}} &       \multicolumn{1}{c}{\textbf{Frosted}}&       \multicolumn{1}{c}{\textbf{Polyethylene}}&       \multicolumn{1}{c}{\textbf{Rough}}&       \multicolumn{1}{c}{\textbf{Sandpaper}}&       \multicolumn{1}{c}{\textbf{Sandpaper}}&       \multicolumn{1}{c}{\textbf{Scouring}}
  & \multicolumn{1}{c}{\textbf{Velcro}}&       \multicolumn{1}{c}{\textbf{Velcro}}&       \multicolumn{1}{c}{\textbf{Wax}}&       \multicolumn{1}{c}{\textbf{White}}\\
\multicolumn{1}{c}{\textbf{}} &   
\multicolumn{1}{c}{\textbf{2}} &       \multicolumn{1}{c}{\textbf{3}} &       \multicolumn{1}{c}{\textbf{acrylic}}&       \multicolumn{1}{c}{\textbf{foam}}&       \multicolumn{1}{c}{\textbf{plastic}}&       \multicolumn{1}{c}{\textbf{220}}&       \multicolumn{1}{c}{\textbf{320}}& 
\multicolumn{1}{c}{\textbf{pad}}&
\multicolumn{1}{c}{\textbf{hooks}}
  &       \multicolumn{1}{c}{\textbf{loops}}&       \multicolumn{1}{c}{\textbf{paper}}&
\multicolumn{1}{c}{\textbf{board}}
\\
\midrule
\midrule
Pair 1
&82e-2 
&64e-2
&13e-2
&51e-2
&26e-2
&21e-2
&75e-2
&37e-2
&27e-2
&98e-2
&97e-2
&73e-2
\\
\midrule
Pair 2
&\colorbox{black!22}{1.5e-5} 
&\colorbox{black!22}{5.8e-5}
&\colorbox{black!22}{7.8e-4}
&\colorbox{black!22}{1.2e-7}
&3.1e-2
&\colorbox{black!22}{4.4e-4}
&\colorbox{black!22}{1.1e-4}
&\colorbox{black!22}{6.0e-8}
&\colorbox{black!22}{2.9e-5}
&\colorbox{black!22}{3.0e-6}
&2.4e-3
&2.4e-3\\

&\colorbox{black!22}{4.2e-2} 
&\colorbox{black!22}{1.1e-1}
&\colorbox{black!22}{1.5e-1}
&\colorbox{black!22}{1.6e-2}
&
&\colorbox{black!22}{1.9e-1}
&\colorbox{black!22}{1.2e-1}
&\colorbox{black!22}{4.3e-2}
&\colorbox{black!22}{1.3e-1}
&\colorbox{black!22}{6.6e-2}
&
&
\end{tabular}}
\caption{Significance p-values for post-hoc two sided Wilcoxon tests for similarity ratings portion of the study. Pair 1: (Real vs.\ NN compared to Real vs.\ AR), Pair 2: (Real vs.\ NN compared to AR vs.\ NN). Statistically significant pairs using Bonferroni corrected cut-off are highlighted and their common language effect sizes are shown below their p-values.}
\label{pvaluetable}
\end{table*}

\begin{figure}[hb]
\begin{center}
\includegraphics[width=1.0\linewidth]{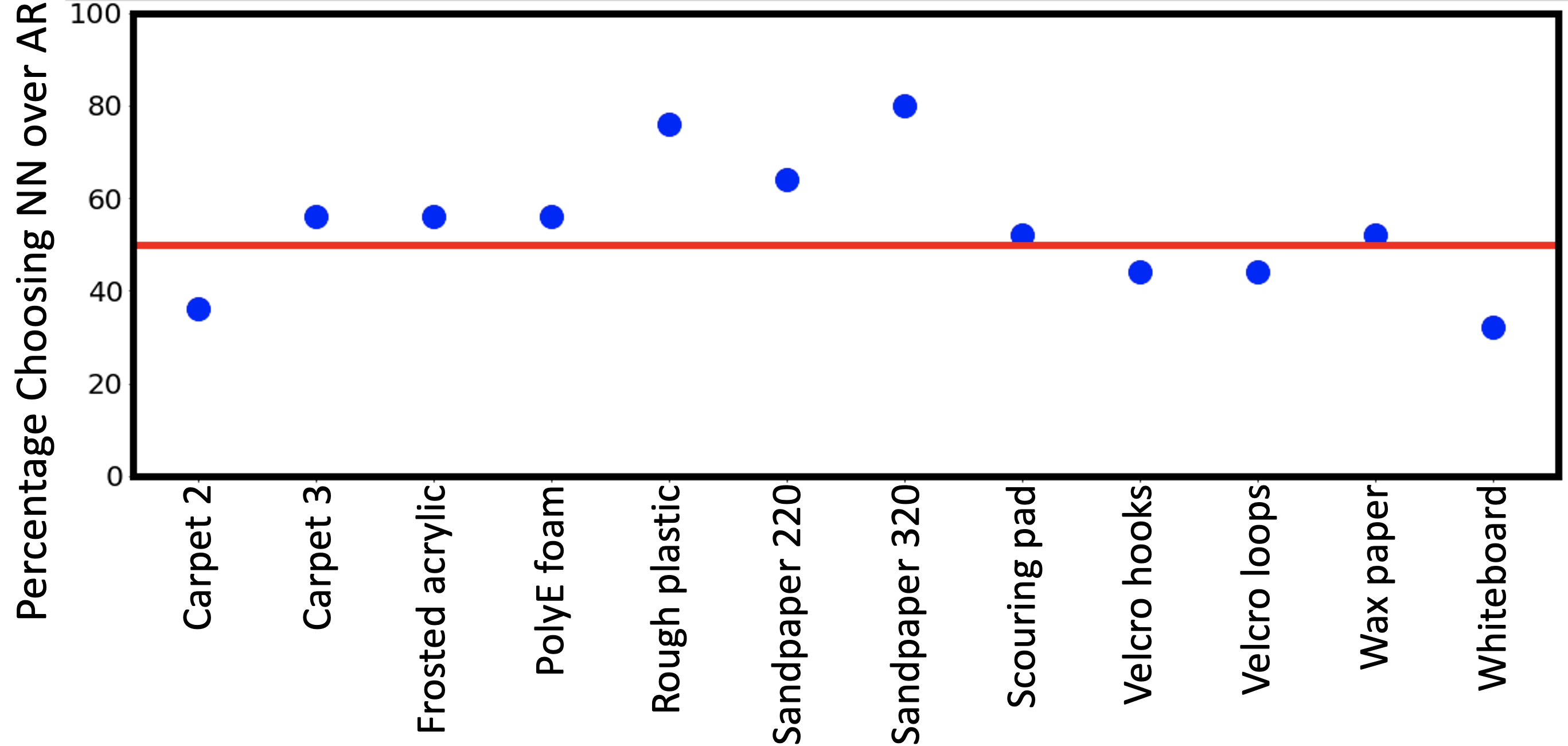}
\end{center}
\caption{Results of the forced-choice comparison experiments. The plot shows the percentage of times our method (NN) was chosen as closer to the real material compared to the piece-wise AR baseline (AR). Red line shows chance performance. NN was chosen as closer to the piece-wise AR baseline for 8 out of 12 materials and statistically outperformed the piece-wise AR baseline for 2 materials. }

\label{fig:forcedchoice}
\end{figure}
Figure~\ref{fig:similarityexp} furthermore shows that the two virtual methods are closer to each other than each virtual method is to the real material (AR vs.\ NN ratings compared to Real vs.\ NN ratings). This is not surprising as there are other aspects of the real textures such as friction and stiffness variations that are not accounted for in these virtual models. Furthermore, the 3D Touch device has a maximum force rendering capability of 3.3 N, while in the original HaTT dataset, participants explored real textures with forces as high as 7 N. This difference in the exerted forces on the participant during virtual exploration versus real texture exploration can decrease sensation similarity. Using a significance value of 0.002, we found the distributions of similarity ratings of Real vs.\ NN to be statistically significantly different than those of AR vs.\ NN for all materials except Rough Plastic, Wax paper, Whiteboard (Table \ref{pvaluetable} row 2). The participants rated this task as easy (2.4 difficulty score out of 5) on average.

\begin{figure*}[ht]
\begin{center}
\includegraphics[width=1.0\linewidth]{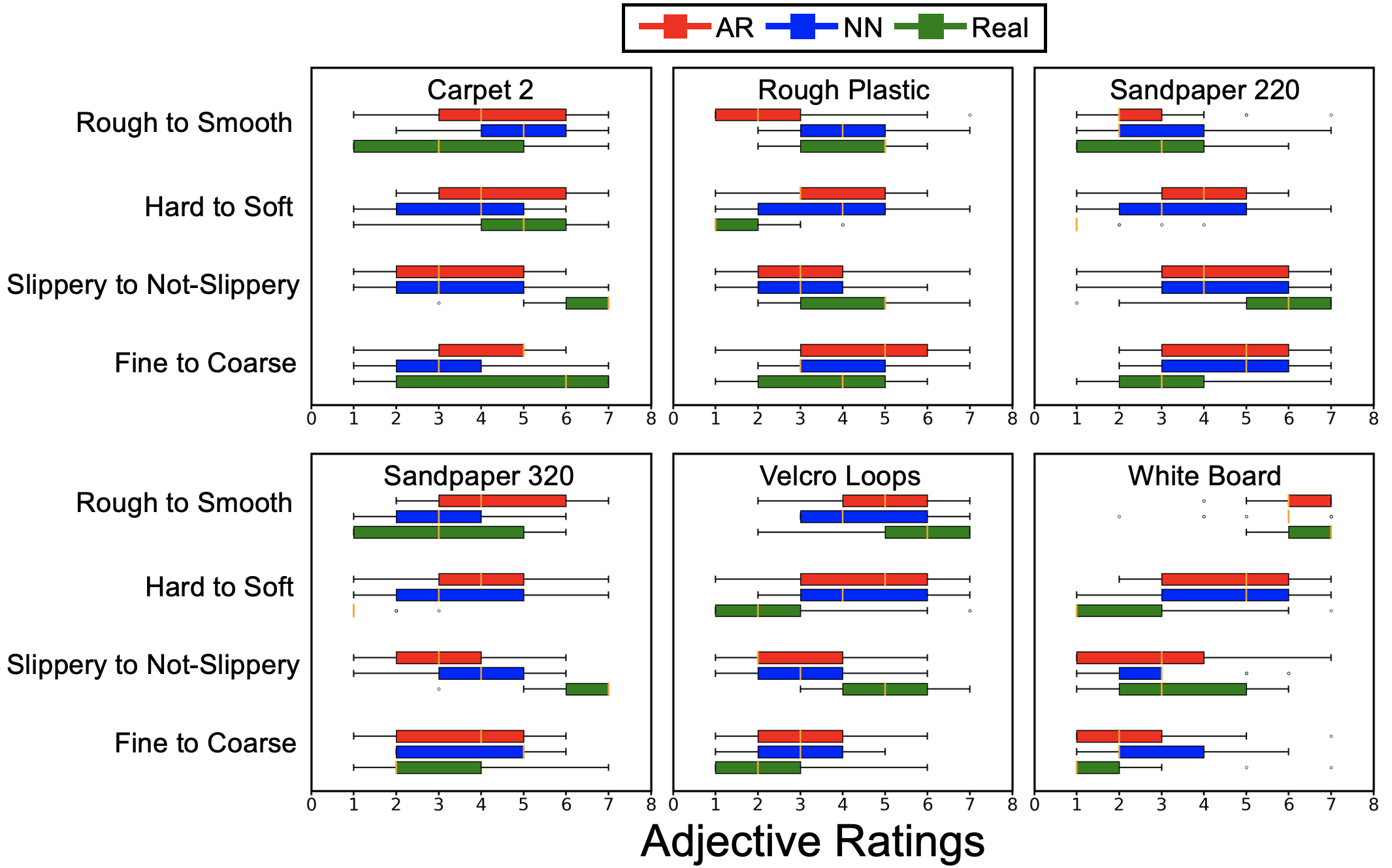}
\end{center}
\caption{Adjective rating experiment results:  Our model was found to outperform the piece-wise AR baseline for rendering Rough plastic and under-performing for Whiteboard. For all the other materials, across both the roughness and coarseness axes, it was observed that both models are performing equally close to the real material. As all the virtual models were only capturing the texture vibration, it is expected that the hardness and slipperiness ratings do not necessarily match that of the real materials for both of these models. These extra adjective axes are included mainly for completeness purposes. The boxes show the range from the first to the third quartile of the ratings across all participants. The orange lines indicate the medians. The whiskers extend the box by 1.5 of the inter-quartile range (IQR) as defined by Tukey's definition of boxplots.}

\label{fig:adjective}
\end{figure*}
\begin{table*}[bt]
\centering
\footnotesize
\scalebox{0.9}{\begin{tabular}{c|ccc|ccc}
\multicolumn{1}{c|}{\textbf{}} &
\multicolumn{3}{c|}{\textbf{Rough-smooth axis}}&
\multicolumn{3}{c}{\textbf{Fine-coarse axis}}
\\
\midrule
\multicolumn{1}{c|}{\textbf{}} & 
\multicolumn{1}{c}{\textbf{AR vs.\ NN}} & 
\multicolumn{1}{c}{\textbf{Real vs.\ AR}} &
\multicolumn{1}{c|}{\textbf{Real vs.\ NN}}&
\multicolumn{1}{c}{\textbf{AR vs.\ NN}} & 
\multicolumn{1}{c}{\textbf{Real vs.\ AR}}&
\multicolumn{1}{c}{\textbf{Real vs.\ NN}}
\\
\midrule
\midrule
Carpet 2
&*
&*
&*
&*
&*
&*
\\
\midrule
Rough Plastic
&\colorbox{black!22}{1.1e-3, 2.4e-1} 
&\colorbox{black!22}{3.0e-3, 7.9e-1}
&{4.5e-1}
&\colorbox{black!22}{3.7e-3, 6.8e-1}
&3.0e-2
&9.8e-1

\\
\midrule
Sandpaper 220
&*
&*
&*
&8.5e-1
&\colorbox{black!22}{9.3e-4, 2.4e-1}
&\colorbox{black!22}{1.8e-3, 2.5e-1}
\\
\midrule
Sandpaper 320
&*
&*
&*
&*
&*
&*
\\
\midrule
Velcro Loops
&2.5e-1
&1.5e-1
&1.9e-2
&3.4e-1
&6.1e-2
&1.4e-1
\\
\midrule
White Board
&*
&*
&*
&1.6e-1
&1.0e-2
&\colorbox{black!22}{2.6e-4, 2.3e-1}

\\
\end{tabular}}
\caption{Significance p-values for post-hoc two sided Wilcoxon tests for adjective ratings portion of the study. Statistically significant pairs using Bonferroni corrected cut-off are highlighted and their p-values are followed by their common language effect sizes. * indicates that a repeated measure Friedman test did not detect statistical significance for that material hence a post-hoc test was not conducted.}
\label{adjectivepvaluetable}
\end{table*}
\subsubsection{Forced Choice Comparison}

The results for this part of the experiment are shown in Figure \ref{fig:forcedchoice}, where we plot the percentage of times our model was chosen over the piece-wise AR baseline model. When our method and the baseline are performing equally well in texture rendering, we expect a pure chance value of 50\%. When our model is outperforming the baseline, we expect to see a percentage higher than 50\%. We observed that our NN-based model was on average preferred for 8 out of the 12 materials, and it was significantly better for 2 materials (Rough plastic (p = 0.007) and sandpaper 320 (p = 0.002)) using a one-sided binomial test with p-value of 0.05. This suggests that our model performs better than the baseline in rendering rough, stiff materials and not statistically significantly different for other materials. 
On average, the participants rated this task as moderately difficult (difficulty score of 3.04 out of 5) which is expected because making a forced choice when the models are performing equally well can be hard.

\subsubsection{Adjective Ratings}
The results of the adjective rating portion of the study is shown in Figure \ref{fig:adjective}. In this study, we would like our model's adjective ratings to be close to that of the real material. An ideal model would not show any statistically significant difference with the real material here. We conducted one-way repeated measure Friedman tests separately for each texture and each adjective using the model type as factor (Real, AR, and NN) followed by post-hoc two-sided Wilcoxon tests with continuity and Bonferroni corrections where appropriate. Table \ref{adjectivepvaluetable} shows the statistical results for rough-smooth and fine-coarse axes, and we discuss the results of the other two axes in text because they were the same for the two virtual rendering methods. We state that our model is likely to be performing better than the piece-wise AR baseline along an adjective axis for a material, when our model's ratings are not statistically significantly different from the real material while the AR baseline's ratings are statistically significantly different. When both our model and the piece-wise AR baseline fail to reject the null hypothesis, we cannot conclude that a difference exists between either of the models and the real material. When both our model and the piece-wise AR baseline are statically significantly different from the real material, we conclude both models are failing to capture the real material properly along that axis. Finally, we conclude that our model is likely to be performing worse than the piece-wise AR baseline along an adjective axis for a material when our model's ratings are statistically significantly different from the real material while the AR baseline's ratings are not statistically significantly different.

We make the following observations (Table \ref{adjectivepvaluetable}, Figure \ref{fig:adjective}): 
\begin{itemize}

\item Rough-smooth axis: 
The Friedman test did not show significance of model type for any of the textures except for Rough Plastic (p = 0.001) and Velcro Loops (p = 0.006) across this axis. As a result, except for these two materials, we cannot conclude that a statistically significant difference exists between either of the models and the real material when modeling roughness/smoothness. Post-hoc two-sided Wilcoxon tests with continuity and Bonferroni correction (significance = 0.017) suggested that there is a statistically significant difference between the Real vs.\ AR and AR vs.\ NN pairs, while no significance was observed for the Real vs.\ NN pair for Rough Plastic. Visually, it can also be seen in Figure \ref{fig:adjective} that our model has a closer median rating to the real material compared to the baseline AR model for Rough plastic. This is in line with our observation in the previous experiments where our model was outperforming the piece-wise AR baseline for rough textures. For Velcro Loops, none of the pairs showed any statistical significance.

\item Hard-soft axis: 
The Friedman test showed significance of model type for all the textures except for Carpet across this axis. For all the significant textures except Velcro Loops, post-hoc Wilcoxon test showed that a significant difference exists between both the Real vs.\ AR and Real vs.\ NN pairs but not between the AR vs.\ NN pairs. This was expected as we were only modeling texture vibrations in these experiments, and the virtual hardness was set to a constant value across all virtual textures regardless of the surface properties. This is in line with previous studies that showed high-frequency vibration modelings are not sufficient to capture hardness and slipperiness properties \cite{Culbertson2014ModelingAR}. Using a significant threshold of 0.003 (Bonferroni correction for 15 paired test), Velcro Loops did not show a significant difference between AR vs.\ NN (p = 0.53), Real vs.\ NN (p = 0.006), and Real vs.\ AR (p = 0.009). However, it should be noted that Bonferroni is a rather conservative threshold.

\item Slippery-not-slippery axis: 
All materials except for Rough plastic and White board had significance of model type with Friedman testing. Using post-hoc Wilcoxon tests with corrections (significance level 0.004), we found Carpet 2, Sandpaper 320, and Velcro loops to have significant differences for Real vs.\ AR and Real vs.\ NN pairs but not for the AR vs.\ NN pair. For Sandpaper 220, no model pairs showed statistical significance. This was expected as the friction was held constant across both models and high-frequency vibrations are not enough to convey the frictional properties of the surfaces as also seen in previous studies \cite{Culbertson2014ModelingAR}.

\item Fine-coarse axis: 
Rough plastic, Sandpaper 220, Velcro Loops, and White board showed statistical significance of model type with Friedman testing. Among these, Rough plastic only showed significance for the AR vs.\ NN pair during post-hoc Wilcoxon test using a Bonferroni correction (significance threshold of 0.004). This means a significant difference exists between fine-coarseness perception of our model and the baseline. This information along with the observation in the forced-choice experiments, where users chose our model over the baseline, suggests that our model is outperforming the baseline for Rough plastic for coarseness. 
Both Sandpaper 220 and Whiteboard showed significance for Real vs.\ NN pairs, but only Sandpaper 220 showed significance for the Real vs.\ AR pair. Neither of the materials showed significance for the AR vs.\ NN pairs. This suggests both models failed to capture this adjective for Sandpaper 220, and the AR model is performing better for Whiteboard. This is inline with our observation in the forced-choice experiments where less than half of the participants chose our model over the AR baseline for Whiteboard. 
\end{itemize}

Overall, we observed that along the axis relevant to texture vibrations, our model outperformed the AR baseline for Rough plastic, under-performed for Whiteboard for fine-coarseness and did not show significant difference for other materials. This is in line with the observation in the forced-choice comparison experiment. The participants rated this task as moderate (3.16 difficulty score out of 5) on average.

\begin{figure}[ht!]
\begin{center}
\includegraphics[width=1.0\linewidth]{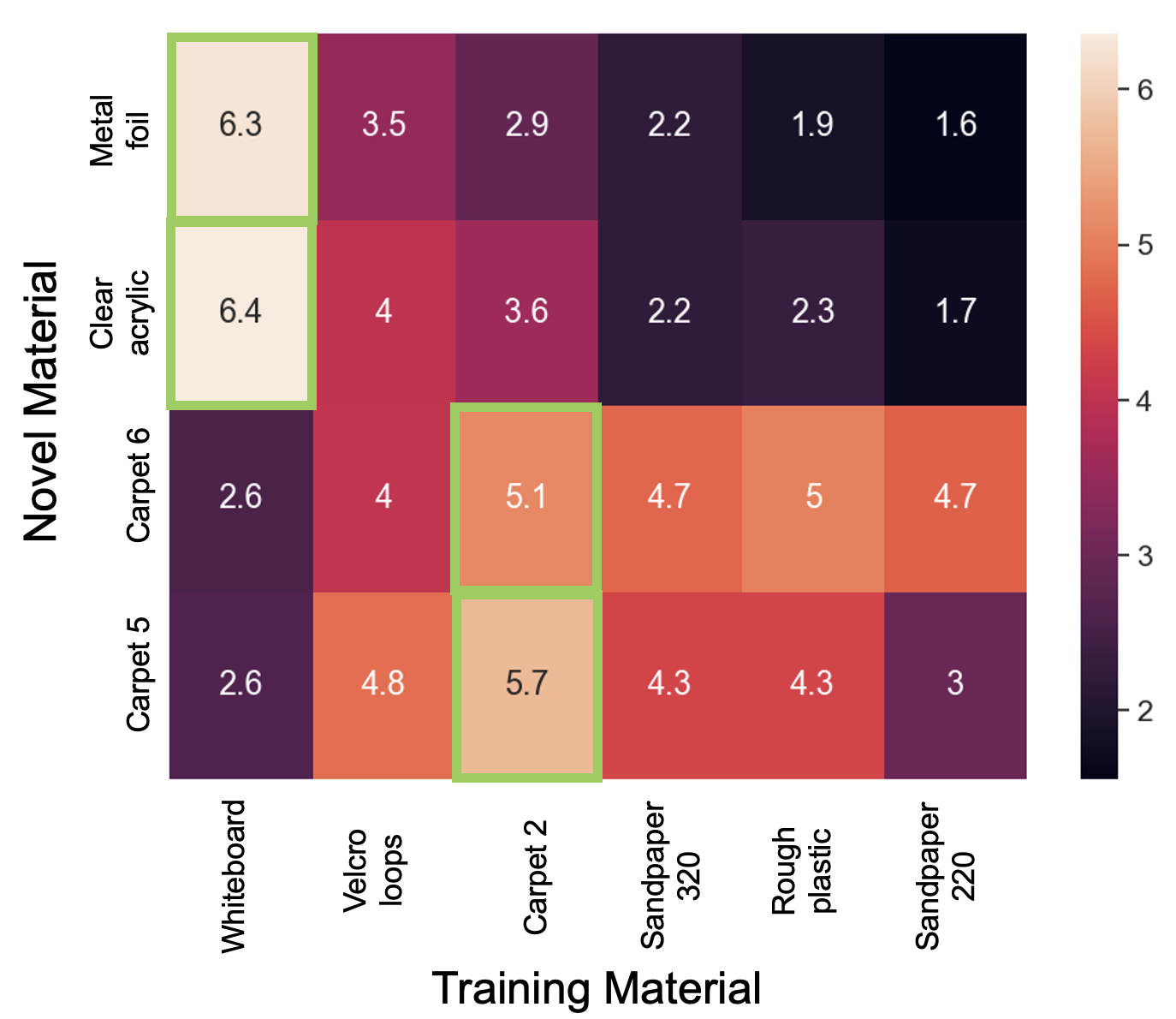}
\end{center}
\caption{Similarity ratings between novel materials and a set of training set materials. Novel material are rated most similar to materials of the same class or similar texture in training (outlined in green). This suggests our model can be used to render novel materials based on their GelSight image only.}

\label{fig:novelmaterial}
\end{figure}

\subsubsection{Evaluation of Unseen Textures}

Figure \ref{fig:novelmaterial} shows that smooth novel materials were rated most similar to the White board model, and the novel carpets were rated most similar to an existing carpet model in the training set. Though carpets were closely matched with the corresponding carpet in the training material set, because they have a rough surface, they were also rated perceptually close to Rough plastic, Sandpaper 320, and Velcro loops. This suggests that our model can be used to render novel materials based on their GelSight images without the need for any further data collection which is not possible in prior state-of-the-art methods. This enables models capable of rendering a wide range of material without the need for an extensive data collection. The participants rated this task as easy (difficulty score of 2 out of 5) on average.

In terms of exploration strategies, participants reported moving the handle back and forth in multiple directions for example both vertically and horizontally, as well as in circular motion or zig-zag patterns, and applying varying force and speeds. Some also reported imagining and assigning a known surface from their memory to the touch sensation they felt.

\section{Conclusions}

In this paper, we built on our prior work \cite{HeraviTexture} to develop a neural network-based model for haptic texture rendering in real time. Given a GelSight image of a surface and human action as input, our model predicts the induced acceleration signal that can be rendered on a haptic device. Because our model is trained using a surface image as input, it is material-aware and unified over all the available textures in the dataset. This reduces the need for developing and maintaining a separate model for each surface. We evaluated the performance of this model in a human user study. The study showed that  our method provided comparable or better performance in modeling texture vibrations without the need for a separate model per texture compared to the state-of-the-art. Because our model is unified over all materials and distinguishes between different textures using their GelSight image, it enables rendering of novel materials that were not in the training set using only a GelSight image of the surface. Our study found that novel smooth materials rendered using this way were rated most similar to existing virtual smooth materials in the training set. Similarly, novel textured materials such as carpet were rated close to other carpets. This suggests that our model is capable of rendering novel materials without the need for an extensive data collection on each surface. In future work, run time of our model can be significantly reduced by optimizing the haptic API code (OpenHaptics \cite{OpenHaptics}) to run on a GPU and faster communication with the Libtorch library \cite{Pytorch}. This will further improve our model's perceptual performance. Furthermore, for rendering of novel materials, we used a nearest neighbor representation instead of the raw representation vector for improved stability of the generated signal. In future work, we can explore using a larger dataset of textures for training to further improve the representation space such that the raw vector for novel textures can be directly used as input to the acceleration generator module.
 
\section*{Acknowledgements}
This research was supported in part by a National Science Foundation Graduate Research Fellowship, seed grants from the Stanford Institute for the Human-Centered Artificial Intelligence (HAI), and an Amazon Research Award. This article solely reflects the opinions and conclusions of its authors and not of Amazon or any entity associated with Amazon.com. We thank Katherine Kuchenbecker and Yasemin Vardar for giving us access to the textures used in the Penn Haptic Texture Toolkit (HaTT) for our GelSight data collection. We also thank Zonghe Chua and Wenzhen Yuan for fruitful discussions. 
\bibliographystyle{plain}
\bibliography{references.bib}

\begin{thebibliography}{10}

\bibitem{SPSI}
G.~Beauregard, M.~Harish, and L.~Wyse.
\newblock Single pass spectrogram inversion.
\newblock In {\em IEEE International Conference on Digital Signal Processing (DSP)}, pages 427--431, 2015.

\bibitem{BohgHSBKSS16}
J.~{Bohg}, K.~{Hausman}, B.~{Sankaran}, O.~{Brock}, D.~{Kragic}, S.~{Schaal}, and G.~S. {Sukhatme}.
\newblock Interactive perception: Leveraging action in perception and perception in action.
\newblock {\em IEEE Transactions on Robotics}, 33(6):1273--1291, 2017.

\bibitem{Burka}
A.~{Burka}, S.~{Hu}, S.~{Helgeson}, S.~{Krishnan}, Y.~{Gao}, L.~A. {Hendricks}, T.~{Darrell}, and K.~J. {Kuchenbecker}.
\newblock Proton: A visuo-haptic data acquisition system for robotic learning of surface properties.
\newblock In {\em IEEE International Conference on Multisensor Fusion and Integration for Intelligent Systems (MFI)}, pages 58--65, 2016.

\bibitem{CipressoVRVisual}
P.~Cipresso, I.~Giglioli, M.~Raya, and G.~Riva.
\newblock The past, present, and future of virtual and augmented reality research: A network and cluster analysis of the literature.
\newblock {\em Frontiers in Psychology}, 9:2086, 2018.

\bibitem{HeatherDoesForceMatter}
H.~Culbertson and K.~J. Kuchenbecker.
\newblock Should haptic texture vibrations respond to user force and speed?
\newblock In {\em IEEE World Haptics Conference (WHC)}, pages 106--112, 2015.

\bibitem{Culbertson2014OneHD}
H.~{Culbertson}, J.~{Lopez Delgado}, and K.~J. {Kuchenbecker}.
\newblock One hundred data-driven haptic texture models and open-source methods for rendering on {3D} objects.
\newblock {\em IEEE Haptics Symposium (HAPTICS)}, pages 319--325, 2014.

\bibitem{RefinedARMA}
H.~{Culbertson}, J.~M. {Romano}, P.~{Castillo}, M.~{Mintz}, and K.~J. {Kuchenbecker}.
\newblock Refined methods for creating realistic haptic virtual textures from tool-mediated contact acceleration data.
\newblock In {\em IEEE Haptics Symposium (HAPTICS)}, pages 385--391, 2012.

\bibitem{Culbertson2014ModelingAR}
H.~Culbertson, J.~Unwin, and K.~J. Kuchenbecker.
\newblock Modeling and rendering realistic textures from unconstrained tool-surface interactions.
\newblock {\em IEEE Transactions on Haptics}, 7(3):381--393, 2014.

\bibitem{GANs}
I.~Goodfellow, J.~Pouget-Abadie, M.~Mirza, B.~Xu, D.~Warde-Farley, S.~Ozair, A.~Courville, and Y.~Bengio.
\newblock Generative adversarial networks.
\newblock {\em Communication of the ACM}, 63(11):139–144, 2020.

\bibitem{GLA}
D.~{Griffin} and J.~{Lim}.
\newblock Signal estimation from modified short-time fourier transform.
\newblock {\em IEEE Transactions on Acoustics, Speech, and Signal Processing}, 32(2):236--243, 1984.

\bibitem{HeraviTexture}
N.~Heravi, W.~Yuan, A.~M. Okamura, and J.~Bohg.
\newblock Learning an action-conditional model for haptic texture generation.
\newblock In {\em IEEE International Conference on Robotics and Automation (ICRA)}, pages 11088--11095, 2020.

\bibitem{OpenHaptics}
B.~Itkowitz, J.~Handley, and W.~Zhu.
\newblock {The OpenHaptics/spl trade/ toolkit: a library for adding 3D Touch/spl trade/ navigation and haptics to graphics applications}.
\newblock In {\em First Joint Eurohaptics Conference and Symposium on Haptic Interfaces for Virtual Environment and Teleoperator Systems. World Haptics Conference}, pages 590--591, 2005.

\bibitem{GelSight1}
M.~K. {Johnson} and E.~H. {Adelson}.
\newblock Retrographic sensing for the measurement of surface texture and shape.
\newblock In {\em IEEE Conference on Computer Vision and Pattern Recognition}, pages 1070--1077, 2009.

\bibitem{GelSight2}
M.~K. Johnson, F.~Cole, A.~Raj, and E.~H. Adelson.
\newblock Microgeometry capture using an elastomeric sensor.
\newblock {\em ACM Transactions on Graphics}, 30(4):46:1--46:8, 2011.

\bibitem{SpationTemporalTextureRendering}
J.~B. Joolee and S.~Jeon.
\newblock Data-driven haptic texture modeling and rendering based on deep spatio-temporal networks.
\newblock {\em IEEE Transactions on Haptics}, 15(1):62--67, 2022.

\bibitem{KlatzkyTextureProbe}
R.~L. Klatzky, S.~J. Lederman, C.~Hamilton, M.~Grindley, and R.~H. Swendsen.
\newblock Feeling textures through a probe: Effects of probe and surface geometry and exploratory factors.
\newblock {\em Perception {\&} Psychophysics}, 65(4):613--631, 2003.

\bibitem{Alexnet}
A.~Krizhevsky, I.~Sutskever, and G.~E.~Hinton.
\newblock Imagenet classification with deep convolutional neural networks.
\newblock {\em Communications of the ACM}, 60(6):84–90, 2017.

\bibitem{Lin2008HapticR}
M.~C. Lin and M.~A. Otaduy.
\newblock {\em Haptic Rendering: Foundations, Algorithms, and Applications}.
\newblock CRC Press, 2008.

\bibitem{ShihanHeatherPrefernce}
S.~Lu, M.~Zheng, M.~C. Fontaine, S.~Nikolaidis, and H.~M. Culbertson.
\newblock Preference-driven texture modeling through interactive generation and search.
\newblock {\em IEEE Transactions on Haptics}, 15(3):508--520, 2022.

\bibitem{Magnenatbelievability}
N.~Magnenat-Thalmann, H.~Kim, A.~Egges, and S.~Garchery.
\newblock Believability and interaction in virtual worlds.
\newblock In {\em 11th International Multimedia Modelling Conference}, pages 2--9, 2005.

\bibitem{DyanmicModelingHaptuator}
W.~McMahan and K.~J. Kuchenbecker.
\newblock Dynamic modeling and control of voice-coil actuators for high-fidelity display of haptic vibrations.
\newblock In {\em IEEE Haptics Symposium (HAPTICS)}, pages 115--122, 2014.

\bibitem{WaveformTexture}
W.~Nai, J.~Liu, C.~Sun, Q.~Wang, G.~Liu, and X.~Sun.
\newblock Vibrotactile feedback rendering of patterned textures using a waveform segment table method.
\newblock {\em IEEE Transactions on Haptics}, 14(4):849--861, 2021.

\bibitem{AmazonVirtualTryon}
A.~Neuberger, E.~Borenstein, B.~Hilleli, E.~Oks, and S.~Alpert.
\newblock Image based virtual try-on network from unpaired data.
\newblock In {\em IEEE/CVF Conference on Computer Vision and Pattern Recognition (CVPR)}, pages 5183--5192, 2020.

\bibitem{NoeBook}
A.~Noe.
\newblock {\em Action in Perception}.
\newblock The MIT Press, 2004.

\bibitem{Okamoto2013PsychophysicalDO}
S.~Okamoto, H.~Nagano, and Y.~Yamada.
\newblock Psychophysical dimensions of tactile perception of textures.
\newblock {\em IEEE Transactions on Haptics}, 6:81--93, 2013.

\bibitem{Pytorch}
A.~Paszke, S.~Gross, F.~Massa, A.~Lerer, J.~Bradbury, G.~Chanan, T.~Killeen, Z.~Lin, N.~Gimelshein, L.~Antiga, A.~Desmaison, A.~Kopf, E.~Yang, Z.~DeVito, M.~Raison, A.~Tejani, S.~Chilamkurthy, B.~Steiner, L.~Fang, J.~Bai, and S.~Chintala.
\newblock {PyTorch: An Imperative Style, High-Performance Deep Learning Library}.
\newblock In {\em Advances in Neural Information Processing Systems}, pages 8024--8035. Curran Associates, Inc., 2019.

\bibitem{phaselockedvocoders}
M.~Puckette.
\newblock Phase-locked vocoder.
\newblock In {\em Proceedings of Workshop on Applications of Signal Processing to Audio and Accoustics}, pages 222--225, 1995.

\bibitem{Romano}
J.~M. Romano and K.~J. Kuchenbecker.
\newblock Creating realistic virtual textures from contact acceleration data.
\newblock {\em IEEE Transactions on Haptics}, 5(2):109--119, 2012.

\bibitem{ANOVALikert}
M.~L. Schrum, M.~Johnson, M.~Ghuy, and M.~C. Gombolay.
\newblock Four years in review: Statistical practices of likert scales in human-robot interaction studies.
\newblock In {\em Companion of the 2020 ACM/IEEE International Conference on Human-Robot Interaction}, HRI, page 43–52, New York, NY, USA, 2020. Association for Computing Machinery.

\bibitem{FreqDecomposedNNTextureRendering}
S.~Shin, R.~Haghighi~Osgouei, K.~Kim, and S.~Choi.
\newblock Data-driven modeling of isotropic haptic textures using frequency-decomposed neural networks.
\newblock In {\em IEEE World Haptics Conference (WHC)}, pages 131--138, 2015.

\bibitem{ConstrainedLMT}
M.~{Strese}, J.~{Lee}, C.~{Schuwerk}, Q.~{Han}, H.~{Kim}, and E.~{Steinbach}.
\newblock A haptic texture database for tool-mediated texture recognition and classification.
\newblock In {\em IEEE International Symposium on Haptic, Audio and Visual Environments and Games (HAVE) Proceedings}, pages 118--123, 2014.

\bibitem{LMTMulti}
M.~{Strese}, C.~{Schuwerk}, A.~{Iepure}, and E.~{Steinbach}.
\newblock Multimodal feature-based surface material classification.
\newblock {\em IEEE Transactions on Haptics}, 10(2):226--239, 2017.

\bibitem{Ujitoko2018VibrotactileSG}
Y.~Ujitoko and Y.~Ban.
\newblock Vibrotactile signal generation from texture images or attributes using generative adversarial network.
\newblock In {\em Haptics: Science, Technology, and Applications}, pages 25--36. Springer International Publishing, 2018.

\bibitem{Vallat2018}
R.~Vallat.
\newblock Pingouin: statistics in python.
\newblock {\em Journal of Open Source Software}, 3(31):1026, 2018.

\bibitem{2020SciPy-NMeth}
P.~Virtanen, R.~Gommers, T.~E. Oliphant, M.~Haberland, T.~Reddy, D.~Cournapeau, E.~Burovski, P.~Peterson, W.~Weckesser, J.~Bright, S.~J. {van der Walt}, M.~Brett, J.~Wilson, K.~J. Millman, N.~Mayorov, A.~R.~J. Nelson, E.~Jones, R.~Kern, E.~Larson, C.~J. Carey, {\.I}.~Polat, Y.~Feng, E.~W. Moore, J.~{VanderPlas}, D.~Laxalde, J.~Perktold, R.~Cimrman, I.~Henriksen, E.~A. Quintero, C.~R. Harris, A.~M. Archibald, A.~H. Ribeiro, F.~Pedregosa, P.~{van Mulbregt}, and {SciPy 1.0 Contributors}.
\newblock {{SciPy} 1.0: Fundamental Algorithms for Scientific Computing in Python}.
\newblock {\em Nature Methods}, 17:261--272, 2020.

\bibitem{Wang2017TacotronTE}
Y.~Wang, R.~J. Skerry-Ryan, D.~Stanton, Y.~Wu, R.~J. Weiss, N.~Jaitly, Z.~Yang, Y.~Xiao, Z.~Chen, S.~Bengio, Q.~V. Le, Y.~Agiomyrgiannakis, R.~A.~J. Clark, and R.~A. Saurous.
\newblock Tacotron: Towards end-to-end speech synthesis.
\newblock In {\em Interspeech}, 2017.

\end{thebibliography}

\end{document}